\documentclass[10pt,conference]{IEEEtran}

\usepackage{cite}

\ifCLASSINFOpdf
   \usepackage[pdftex]{graphicx}
   \graphicspath{{figs/}}
   \DeclareGraphicsExtensions{.pdf,.jpeg,.png}
\else
   \usepackage[dvips]{graphicx}
   \graphicspath{{../figs/}}
   \DeclareGraphicsExtensions{.eps}
\fi
\usepackage[cmex10]{amsmath}
\usepackage{amsthm}

\usepackage{algorithmic}

\usepackage{array}

\ifCLASSOPTIONcompsoc
  \usepackage[caption=false,font=normalsize,labelfont=sf,textfont=sf]{subfig}
\else
  \usepackage[caption=false,font=footnotesize]{subfig}
\fi
\usepackage{url}
\usepackage{lipsum}
\usepackage{color}
\usepackage{xcolor}
\usepackage{adjustbox}
\usepackage{booktabs}
\usepackage{multirow}
\usepackage{amsfonts}

\newif\iffinal
\finalfalse

\iffinal
\else
\usepackage[switch]{lineno}
\fi

\begin{document}
%
\title{Data Augmentation  Improves Machine Unlearning}




\author{\IEEEauthorblockN{Andreza M. C. Falcao, Filipe R. Cordeiro}
\IEEEauthorblockA{Visual Computing Lab, Department of Computing, 
Universidade Federal Rural  de Pernambuco, Brazil}
  andreza.mcfalcao@ufrpe.br, filipe.rolim@ufrpe.br}

\maketitle

\begin{abstract}
Machine Unlearning (MU) aims to remove the influence of specific data from a trained model while preserving its performance on the remaining data. 
Although a few works suggest connections between memorisation and augmentation, the role of systematic augmentation design in MU remains under-investigated. In this work, we investigate the impact of different data augmentation strategies on the performance of unlearning methods, including SalUn, Random Label, and Fine-Tuning. Experiments conducted on CIFAR-10 and CIFAR-100, under varying forget rates, show that proper augmentation design can significantly improve unlearning effectiveness, reducing the performance gap to retrained models. Results showed a reduction of up to 40.12\% of the Average Gap unlearning Metric, when using TrivialAug augmentation. Our results suggest that augmentation not only helps reduce memorization but also plays a crucial role in achieving privacy-preserving and efficient unlearning.

\end{abstract}


\IEEEpeerreviewmaketitle

\section{Introduction}

Machine Learning (ML) models rely on large volumes of training data, which often contain sensitive, copyrighted or private information~\cite{buick2025copyright}. The construction of a training set of ML models is usually not transparent and may use personal data without permission. For this reason,  privacy concerns have emerged in ML model design, driven by regulations such as the \textit{Right to be Forgotten}~\cite{hoofnagle2019european}, which guarantee individuals the right to request the removal of their personal data from organizational systems~\cite{dang2021right}. 
Privacy regulations such as the General Data Protection Regulation (GDPR)~\cite{mantelero2013eu}, the Personal Information Protection and Electronic Documents Act (PIPEDA)~\cite{jaar2008canadian}, and the California Consumer Privacy Act (CCPA) underscore the importance of protecting personal data.
According to these laws, companies must take reasonable measures to guarantee that personal data is deleted upon request. It indicates that individual users have the right to request companies to remove their private data, which was previously collected for ML model training. In the context of ML models, addressing such requests necessitates not merely the deletion of data from databases but also effectively removing its influence from already trained models. 
This task, known as Machine Unlearning (MU), involves updating models so that they behave as if the specific data had never been part of their training sets. Beyond compliance, MU also enhances model robustness by mitigating adverse effects from harmful or outdated data, such as poisoned examples or noisy labels, and significantly reduces vulnerability to privacy attacks, including membership inference and model inversion~\cite{randomlabel, graves2021amnesiac}.
Traditionally, complying with unlearning requests involves fully retraining models from scratch on the remaining dataset, a process that, while ensures compliance with regulations, is computationally costly and impractical for frequent data removal requests, especially with complex models such as large language models (LLMs)~\cite{shanahan2024talking}. 

\begin{figure}
    \centering
    \includegraphics[width=0.9\linewidth]{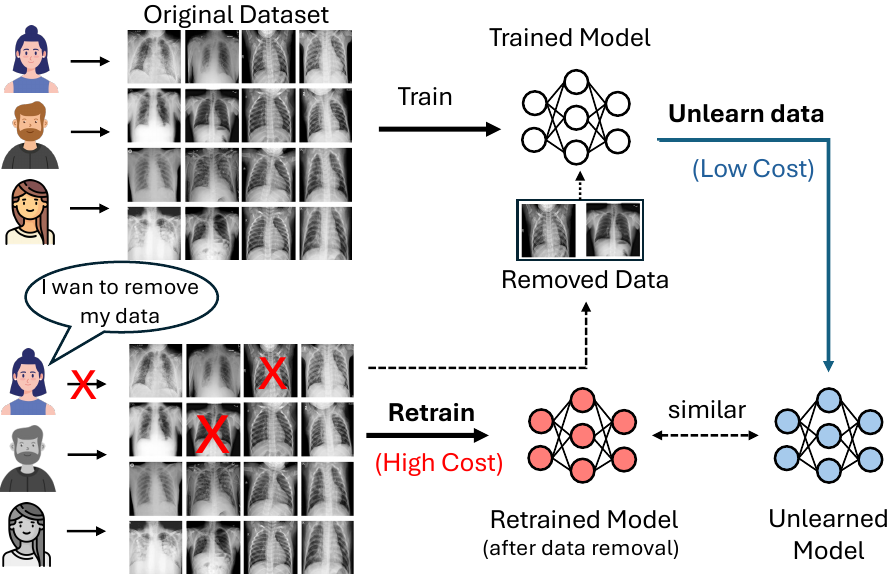}
    \caption{Machine Unlearning process. When a data removal request is made, an unlearned model is generated, free from the influence of the removed samples. The goal of unlearning is to produce a model that behaves similarly to one retrained from scratch but with a much lower computational cost.}
    \label{fig:mu}
\end{figure}

To address this problem, the area of Machine Unlearning has emerged to explore ways to eliminate the influence of specific data from a previously trained model without the need to fully retrain the model without the requested data. 
The main objective of MU is to obtain a model which behaves similarly to one trained from scratch without that data. Figure~\ref{fig:mu} illustrates this process in a medical domain, where the user requests his personal data to be removed from the trained model. Once a user requests the removal of personal data, MU aims to eliminate the influence of that data so that the model behaves as if it had been retrained without the computational cost of retraining.

Several studies have been proposed to investigate the effectiveness of MU methods in image classification tasks~\cite{zhang2023review}. For instance, label smoothing~\cite{di2024label} and network pruning~\cite{jia2023model} have been shown to improve the unlearning process by reducing memorisation and model complexity. Zhao et al.~\cite{zhao2024makes} further demonstrate that memorisation makes unlearning particularly challenging. These works highlight the importance of mitigating memorisation effects in MU. However, while they point to promising directions, they do not systematically explore the role of data augmentation. Since augmentation is a standard tool to control memorisation and biases in machine learning~\cite{mumuni2024survey}, our study extends these preliminary ideas by providing the first comprehensive evaluation of multiple augmentation strategies across different MU methods and forgetting scenarios.

In this work, we investigate the role of data augmentation on machine unlearning techniques for image classification tasks.
Specifically, we analyze the effectiveness of 6 augmentation strategies applied to state-of-the-art (SOTA) MU methods, including Saliency Unlearning (SalUn)~\cite{fan2024salun}, Random Label~\cite{randomlabel}, and Fine-Tuning\cite{finetune}.  We hypothesize that strategic use of data augmentation can significantly reduce memorization, thereby enhancing the efficiency and effectiveness of machine unlearning.
 Our evaluations are conducted on the CIFAR-10 and CIFAR-100~\cite{cifar} datasets across multiple forget rates and unlearning scenarios. 
Results show evidence that the choice of data augmentations improves machine unlearning.





Our main contributions are: (i) a systematic evaluation of 6 augmentation strategies on SalUn, Random Label, and Fine-Tuning; (ii) evidence that augmentation substantially narrows the performance gap to retrained models, especially under high forget rates in CIFAR-100; and (iii) an analysis of class-wise forgetting scenarios, showing the sensitivity of MU methods to augmentation choices.

\section{Prior Work}


Machine Unlearning approaches can be categorized into exact and approximate methods. Exact methods ensure the complete removal of the specific data subset and aim to retrain the model with lower computation cost compared to full retraining~\cite{li2024taxonomy}. Techniques within this class typically leverage checkpoint-based retraining strategies, selectively updating portions of the model rather than performing full retraining~\cite{bourtoule2021machine}.
 Although more computationally efficient than complete retraining, exact methods remain prohibitively costly for frequent or large-scale unlearning tasks.
Approximate methods, on the other hand, aim to reduce the influence of the data on the model but do not guarantee its complete removal. However, this approach is computationally more efficient than exact methods and more suitable for scenarios with a high frequency of unlearning requests or when the cost is too high, such as those involving LLMs. 

Previous studies have attempted to reduce the influence of removed data through different strategies. Golatkar et al.~\cite{golatkar2020eternal} proposed a function that perturbs the model weights, ensuring that any information about the data to be forgotten is eliminated. This is achieved by leveraging the stability properties of stochastic gradient descent (SGD) to estimate and minimize the influence of the targeted data on the model’s parameters.  Warnecke et al.~\cite{warnecke2021machine} leveraged influence functions to estimate data influence on model parameters, subsequently applying analytical updates to minimize this impact. Graves et al.~\cite{graves2021amnesiac} demonstrate that deleting data from the training set is insufficient, as models can still leak information through attacks such as model inversion and membership inference. By applying gradient ascent, the model parameters are adjusted to effectively erase traces of the forgotten data, thereby mitigating privacy risks while maintaining performance on the remaining dataset.

The SalUn method~\cite{fan2024salun} is a SOTA  approximate MU technique that  employs saliency maps to identify critical model parameters influenced by the data subset to be removed. SalUn selectively adjusts these parameters through targeted gradient updates, minimizing reliance on the subset and preserving the model's performance on remaining data. By prioritizing salient parameters, SalUn effectively achieves efficient unlearning with lower computational overhead compared to traditional retraining.


Towards optimising the unlearning process, Jia et al.~\cite{jia2023model} demonstrate that pruning reduces the number of parameters, making MU more efficient. Di et al.~\cite{di2024label} show that employing soft labels alleviates memorisation, leading to improved unlearning outcomes. Zhao et al.~\cite{zhao2024makes} explicitly connect memorisation with unlearning difficulty, highlighting the need for methods that reduce memorisation. These studies indicate the potential value of modifying training strategies to enhance MU. Nevertheless, none of them systematically investigates data augmentation, despite its well-known role in reducing memorisation and improving robustness in standard ML~\cite{mumuni2024survey}. Our work fills this gap by offering the first broad benchmark of modern augmentation strategies in MU, analysing their impact across multiple methods and unlearning scenarios.


Despite these advancements, only limited attention has been given to how data augmentation strategies affect MU. Given that augmentation is ubiquitous in ML training, understanding its systematic role in MU is an open research direction. This work contributes the first comprehensive empirical study evaluating a broad range of modern augmentation strategies across multiple MU methods and forgetting scenarios.

\section{Methodology}

\subsection{Problem Setup}

Let $\mathcal{D}=\{x_i,y_i\}_{i=1}^N$ be a training dataset of images $x_i$ with associated labels $y_i \in \{1,\dots,K\}$, where $K$ is the number of classes and $N$ the total number of samples.
The Machine Unlearning problem can be formally defined as the task of removing the influence of a specific subset of data $\mathcal{D}_f \subset \mathcal{D}$ from  a previously trained model
$\theta_o = \mathcal{A}(\mathcal{D})$, where $\theta_o$ is the set of weights resulting from applying a training algorithm $\mathcal{A}$ to the dataset 
$\mathcal{D}$. Traditional retraining approaches retrain the model on the remaining subset  $\mathcal{D}_r = \mathcal{D}-\mathcal{D}_f$, obtaining the retrained model weights $\theta_r$. This process results in an ideal reference model $\theta_r = A(\mathcal{D}_r)$, retrained from scratch without the data subset $\mathcal{D}_f$.
Given this context, the MU task consists of employing an unlearning algorithm $\mathcal{U}$, which, starting from the trained model $\theta_o$, the subset to be forgotten $\mathcal{D}_f$, and the remaining subset $\mathcal{D}_r$, produces a new model $\theta_u = \mathcal{U}(\theta_o, \mathcal{D}_f, \mathcal{D}_r)$.
It is expected that $\theta_u$ approximates, in terms of output distribution, the ideal retrained model $\theta_r$.

The performance of the unlearned model is evaluated through the metric gap, defined as $MG = |\mathcal{M}_{MU} - \mathcal{M}_{retrain}|$, where $\mathcal{M}_{MU}$ represents the evaluation metric on the unlearned model and $\mathcal{M}_{retrain}$ corresponds to the metric on the retrained model.
The main challenge lies in designing unlearning algorithms that are not only effective at removing the influence of $\mathcal{D}_f$ but also significantly more computationally efficient than full retraining while preserving performance on the remaining dataset $\mathcal{D}_r$.

In this work, we investigate whether the use of data  augmentation function $G(\cdot)$, which generates the augmented set $\mathcal{D}_A = \{(G(x_i), y_i)\}_{i=1}^N$, can enhance the unlearning process by reducing the metric gap $MG$.

\subsection{Data Augmentations Scenarios}

To investigate the influence of data augmentation techniques on machine unlearning, we evaluated 7 distinct augmentation scenarios: NoAug, Default, Default + RandAugment, Default + AutoAugment, Default + Random Erasing, Default + TrivialAug, and Default + AugMix. Each scenario is described below:

\textbf{NoAug}: This scenario refers to training the model without any augmentation. Although it is not a common approach in  MU, it is important to show how augmentation influences the unlearning process.

\textbf{Default}: This refers to using simple Random Crop and Flip augmentations~\cite{takahashi2019data}. We called it default because it is the most common data augmentation for MU~\cite{fan2024salun,randomlabel}.

\textbf{Default + RandAugment}: RandAugment~\cite{randaug} is a data augmentation method that applies a fixed number of randomly selected image transformations with random magnitudes to improve model robustness. In this setup, we combine Random Crop, Flip and RandAugment.

\textbf{Default + AutoAugment}: AutoAugment~\cite{autoaug} is a data augmentation technique that uses a learned policy to select optimal sequences of image transformations for improving model performance. In this setup, we combine Random Crop, Flip and AutoAugment.

\textbf{Default + Random Erasing}: Random Erasing~\cite{randomerasing} is an augmentation technique that randomly removes a rectangular region from an image to improve model robustness to occlusion. In this setup, we combine Random Crop, Flip and Random Erasing.

\textbf{Default + TrivialAug}: TrivialAugment~\cite{trivial} is a lightweight augmentation method that applies simple, computationally cheap transformations like flips, rotations, and colour jitter to improve model generalization. In this setup, we combine Random Crop, Flip and TrivialAugment.

\textbf{Default + AugMix}: AugMix~\cite{augmix} is a data augmentation technique that blends multiple augmented versions of an image to improve model robustness and uncertainty calibration. In this setup, we combine Random Crop, Flip and AugMix.

When we set the data augmentation scenarios, we use it for all training and evaluation pipelines (i.e. baseline, retrained and unlearned models).

\subsection{Machine Unlearning Evaluation}
\label{sec:metrics}

The evaluation metrics for machine unlearning consist of comparing the metrics results of the unlearned model with those obtained with the retrained model, considering a request forget subset data $\mathcal{D}_f$. The retrained approach is retrained from scratch, using only $\mathcal{D}_r=\mathcal{D}-\mathcal{D}_f$, while the MU unlearns from the previoulsy trained model $\theta_o$ (on $\mathcal{D}$), unlearning on $\mathcal{D}_f$ and fine-tuning on $\mathcal{D}_r$. In this work, we adopt the metrics Unlearn Accuracy (UA), Remaining Accuracy (RA), Testing Accuracy (TA), Membership Inference Attack (MIA), Average GAP (AG), and Run-Time Efficiency (RTE), which are also used in recent MU studies~\cite{fan2024salun, di2024label}. Each metric is described below:

\textbf{Unlearning Accuracy (UA)} quantifies the accuracy of the unlearned model on the subset of data $\mathcal{D}_f$. 
Lower UA values indicate successful data forgetting.

\textbf{Remaining Accuracy (RA)} measures the accuracy of the unlearned model on the remaining data $\mathcal{D}_r$. A high RA value indicates that the model preserved its performance on relevant data.

\textbf{Testing Accuracy (TA)} measures the accuracy of the unlearned model on an independent test set $\mathcal{D}_{test}$. A high TA value indicates that the model successfully generalized to unseen data.

\textbf{Membership Inference Attack (MIA)} measures the vulnerability of the unlearned model to membership inference attacks. A low MIA value indicates better protection against such attacks.

\textbf{Average GAP (AG)}  measures how close the unlearned model is to the retrained model. It is calculated as the average of the absolute difference $|MU - Retrain|$ between the UA, RA, TA, and MU metrics of the unlearned model and those of the retrained model.

\textbf{Run-time Efficiency (RTE)} measures the execution time of each method to unlearn. For retraining, it measures the time to retrain the remaining dataset.



\subsection{Dataset}

Experiments were conducted using CIFAR-10 and CIFAR-100 datasets~\cite{cifar}, which are frequently employed for evaluating machine unlearning methods~\cite{fan2024salun,randomlabel}. Both datasets consist of coloured natural images with a resolution of 32×32 pixels. CIFAR-10 contains 60,000 images divided into 10 classes, with 50,000 images for training and 10,000 for testing. CIFAR-100 shares the same overall size but includes 100 fine-grained classes, each containing 600 images (500 for training and 100 for testing), organized into 20 superclasses. 




\subsection{Implementation}

\begin{figure*}[htpb]
    \centering
    \includegraphics[width=0.8\textwidth]{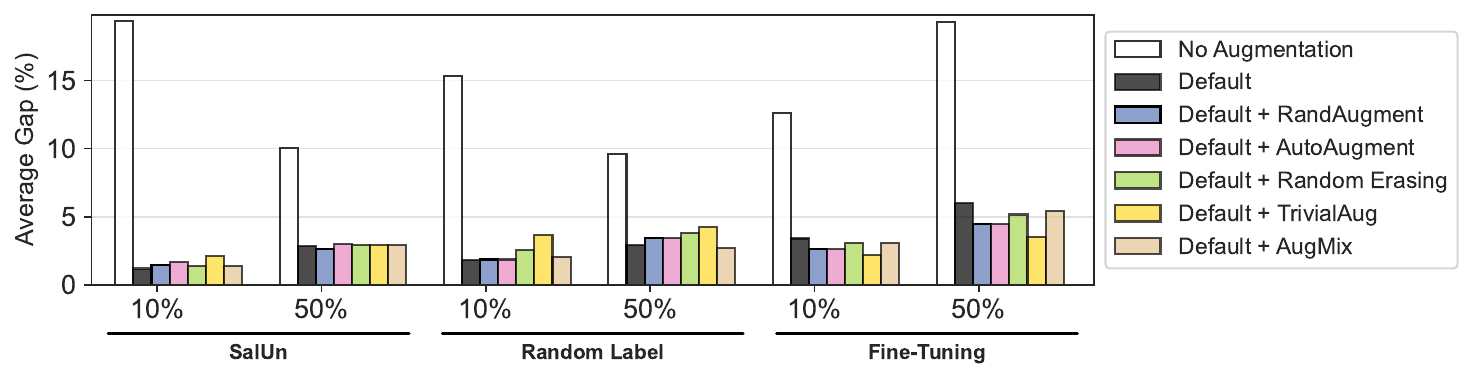}
    \caption{Average Gap (AG) for SalUn, Random Label, and Fine-Tuning on CIFAR-10 with 10\% and 50\% forget rates. Lower AG values indicate better alignment with the retrained model. Fine-Tuning shows the largest improvements with augmentation, especially with TrivialAug.}
    \label{fig:gap_cifar10}
\end{figure*}

\begin{figure*}[htp]
    \centering
    \includegraphics[width=0.8\textwidth]{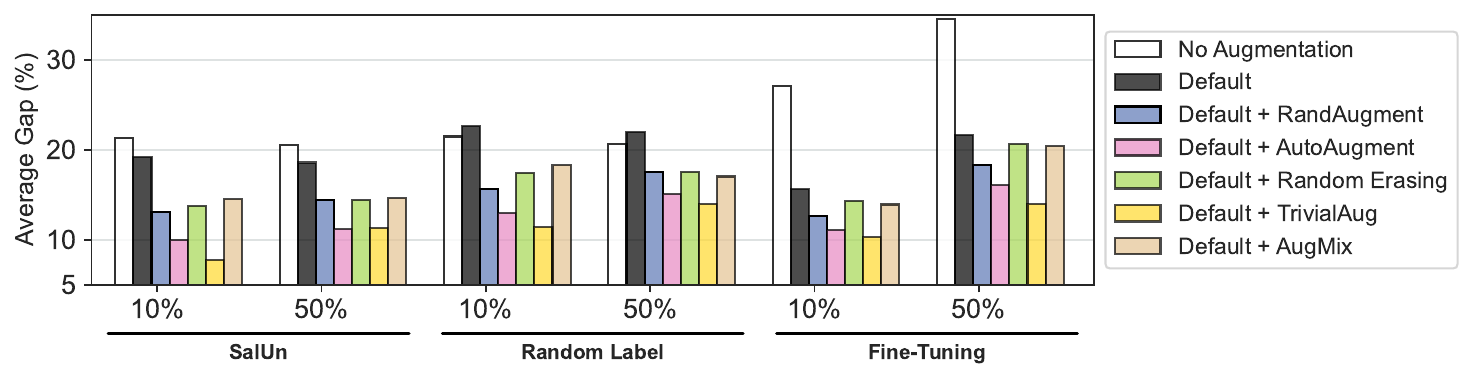}
    \caption{Average Gap (AG) for CIFAR-100 with 10\% and 50\% forget rates. CIFAR-100 exhibits larger gaps due to dataset complexity, but augmentation—particularly TrivialAug—substantially reduces the gap for all methods.}
    \label{fig:gap_cifar100}
\end{figure*}

For baseline model training, we employed the ResNet-18 architecture~\cite{wu2019wider}, trained for 200 epochs, with learning rate of 0.1 and batch size of 256, as in prior studies\cite{fan2024salun}. 
After training the model for 200 epochs, we obtained a trained model $\theta_o$, which is used as the initial condition for applying MU methods. We implemented the MU models SalUn~\cite{fan2024salun}, Random Label~\cite{randomlabel} and Fine-Tuning~\cite{finetune}, following publicly available source codes from previous studies~\cite{fan2024salun}. Each MU method was executed for 10 epochs with a learning rate of 0.01, as used in~\cite{fan2024salun}. From the model $\theta_o$, trained on the entire dataset $\mathcal{D}$, we perform the data forgetting operation, which consists of removing the selected samples from $\mathcal{D}$, resulting in the remaining dataset $\mathcal{D}_r = \mathcal{D} - \mathcal{D}_f$. We simulate the  data request removal by randomly selecting samples from $D$,  using a forgetting rate $\delta_f\in\{10\%, 50\%\}$. This approach is the same as the one used in literature~\cite{fan2024salun, randomlabel}.

To compare with the retrained model, we retrain the model using the same parameters as the baseline, retrained for 200 epochs,  but with the forget samples $D_f$ removed from the dataset. We followed the same procedure used in \cite{fan2024salun}.

 All computational experiments were conducted on a single NVIDIA GeForce RTX 4090 GPU, with an Intel Core i7 processor.
 


\section{Results}

\begin{table*}[!htbp]
\centering
\caption{Comparison of SalUn, Random Label (RL) and Fine-Tuning (FT) with default augmentation and with TrivialAug (TA), for CIFAR-10 and CIFAR-100, with forget rates (FR) 10\% and 50\%. The reported results are given by $a_{\tiny \pm b}$, with mean $a$ and standard deviation $b$ over 5 independent experiments. The average difference between the method results and Retrain (gold model) is presented in parentheses. The better performance of an MU method corresponds to the smaller performance gap with Retrain. The best results for each unlearning method in each scenario are presented in bold. RTE is in minutes. }
\label{tab:metrics}
\scalebox{0.9}{
\begin{tabular}{c|c|c|c|*{5}{c}}
\toprule
\multirow{2}{*}{\textbf{Dataset}}& \multirow{2}{*}{\textbf{FR}} & \multirow{2}{*}{\textbf{Method}} & \multirow{2}{*}{\textbf{Augmentation}}  \\
& & & & $\downarrow$~UA & $\uparrow$~RA & $\uparrow$~TA & $\downarrow$~MIA &  $\downarrow$~RTE  \\
\midrule
\multirow{16}{*}{CIFAR-10} & \multirow{8}{*}{10\%} 
& \multirow{2}{*}{Retrain} & Default & $5.21_{\tiny \pm 0.31} (0.00)$ & $100_{\tiny \pm 0.00}$ (0.00) & $94.44_{\tiny \pm 0.96}$ (0.00) & $13.3_{\tiny \pm 0.36}$ (0.00) & 51.3 \\
  &  &  & Default+TA & $4.34_{\tiny \pm 0.49}$ (0.00)  &  $99.98_{\tiny \pm 0.01}$ (0.00)  & $95.36_{\tiny \pm 0.17}$ (0.00)& $12.07_{\tiny \pm 0.76}$ (0.00) &  50.4 \\
  \cline{3-9}
&  & \multirow{2}{*}{SalUn} & Default & $3.70_{\tiny \pm 0.14} (1.51)$ & $\mathbf{99.03_{\tiny \pm 0.20} (0.97)}$  & $\mathbf{93.27_{\tiny \pm 0.33} (1.17)}$  & $\mathbf{14.31_{\tiny \pm 1.33} (1.21)}$  & 5.4 \\
&  &   & Default+TA & $\mathbf{4.20_{\tiny \pm 0.66} (0.58)}$ & $97.39_{\tiny \pm 0.53}$ (2.59) & $93.20_{\tiny \pm 0.76}$ (2.16)  & $15.35_{\tiny \pm 1.25}$ (3.28) & 5.5 \\
\cline{3-9}
&  & \multirow{2}{*}{RL} & Default & $\mathbf{5.19_{\tiny \pm 0.41} (0.45)}$ & $\mathbf{98.01_{\tiny \pm 0.35} (1.99)}$  & $\mathbf{92.29_{\tiny \pm 0.41} (2.15)}$  &$\mathbf{15.81_{\tiny \pm 1.09} (2.68)}$  & 3.6 \\
&  &  & Default+TA & $6.06_{\tiny \pm 0.99}$ (1.81) & $95.75_{\tiny \pm 0.84}$ (4.23)  & $91.86_{\tiny \pm 0.92}$ (3.51) & $17.10_{\tiny \pm 1.40}$ (5.03)  & 3.6 \\
\cline{3-9}
&  & \multirow{2}{*}{FT} & Default & $1.52_{\tiny \pm 0.24}$ (3.69)  & $\mathbf{99.58_{\tiny \pm 0.11} (0.42)}$ & $\mathbf{93.51_{\tiny \pm 0.16} (0.93)}$ &  $4.51_{\tiny \pm 0.36}$ (8.62) & 4.2 \\
&  &  & Default+TA & $\mathbf{2.50_{\tiny \pm 0.27} (1.82)}$ & $98.87_{\tiny \pm 0.16}$ (1.11) & $93.84_{\tiny \pm 0.37}$ (1.53) &  $\mathbf{7.73_{\tiny \pm 0.55} (4.34)}$ & 4.2 \\
\cline{2-9}
  & \multirow{8}{*}{50\%}  & \multirow{2}{*}{Retrain} & Default & $7.84_{\tiny \pm 0.18}$ (0.00) & $100_{\tiny \pm 0.00}$ (0.00) & $91.76_{\tiny \pm 0.32}$ (0.00) & $19.64_{\tiny \pm 0.39}$ (0.00) & 51.3 \\
  &  &  & Default+TA & $6.39_{\tiny \pm 0.28}$ (0.00) & $99.98_{\tiny \pm 0.01}$ (0.00)  & $93.5_{\tiny \pm 0.35}$ (0.00) & $16.56_{\tiny \pm 0.41}$ (0.00) & 51.4 \\
  \cline{3-9}
&  & \multirow{2}{*}{SalUn} & Default & $6.33_{\tiny \pm 0.76}$ (1.51) & $\mathbf{95.71_{\tiny \pm 0.93} (4.29)}$ & $\mathbf{90.31_{\tiny \pm 0.61} (1.44)}$ & $15.47_{\tiny \pm 1.05}$ (4.17) & 6.0 \\
&  &  & Default+TA & $\mathbf{7.13_{\tiny \pm 1.40} (1.00)}$ & $93.91_{\tiny \pm 1.37}$ (6.08) & $90.01_{\tiny \pm 1.30}$ (3.48) & $\mathbf{15.51_{\tiny \pm 0.69} (1.05)}$ & 5.5 \\
\cline{3-9}
&  & \multirow{2}{*}{RL} & Default & $\mathbf{8.87_{\tiny \pm 1.27} (1.10)}$  & $\mathbf{94.27_{\tiny \pm 1.26} (5.73)}$  & $\mathbf{89.31_{\tiny \pm 0.79} (2.44)}$  & $17.11_{\tiny \pm 1.45}$ (2.53) & 3.6 \\
&  &  & Default+TA & $8.95_{\tiny \pm 1.48}$ (2.56) & $92.25_{\tiny \pm 1.45}$ (7.73) & $88.64_{\tiny \pm 1.30}$ (4.85) & $\mathbf{18.41_{\tiny \pm 0.97} (1.84)}$ & 3.6 \\
\cline{3-9}
&  & \multirow{2}{*}{FT} & Default & $1.17_{\tiny \pm 0.17}$ (6.68) & $99.79_{\tiny \pm 0.07}$ (0.21) & $93.46_{\tiny \pm 0.26}$ (0.71) & $4.15_{\tiny \pm 0.42}$ (15.49) & 2.8 \\
&  &  & Default+TA & $\mathbf{2.32_{\tiny \pm 0.21} (4.06)}$ & $\mathbf{99.12_{\tiny \pm 0.25} (0.86)}$ & $\mathbf{93.66_{\tiny \pm 0.21} (0.38)}$ & $\mathbf{7.72_{\tiny \pm 0.49} (8.84)}$ & 2.3 \\
\midrule

\multirow{16}{*}{CIFAR-100} & \multirow{8}{*}{10\%}   & \multirow{2}{*}{Retrain} & Default & $25.32_{\tiny \pm 0.70}$ (0.00) & $99.98_{\tiny \pm 0.00}$ (0.00) & $74.42_{\tiny \pm 0.38}$ (0.00) & $50.90_{\tiny \pm 0.96}$ (0.00) & 83.3 \\
  &  &  & Default+TA & $22.67_{\tiny \pm 0.57}$ (0.00) & $99.93_{\tiny \pm 0.02}$ (0.00) & $76.91_{\tiny \pm 0.14}$ (0.00) & $44.89_{\tiny \pm 0.56}$ (0.00) & 87.1 \\
  \cline{3-9}
&  & \multirow{2}{*}{SalUn} & Default & $55.18_{\tiny \pm 1.36}$ (29.86) & $\mathbf{99.05_{\tiny \pm 0.14} (0.93)}$ & $67.51_{\tiny \pm 0.75}$ (6.91) & $90.00_{\tiny \pm 0.88}$ (39.09) & 5.2 \\
&  &  & Default+TA & $\mathbf{27.37_{\tiny \pm 2.78} (4.70)}$ &$96.81_{\tiny \pm 0.53}$ (3.12)  & $\mathbf{71.73_{\tiny \pm 0.90} (5.17)}$ & $\mathbf{63.13_{\tiny \pm 1.45} (18.25)}$ & 5.5 \\
\cline{3-9}
&  & \multirow{2}{*}{RL} & Default & $65.80_{\tiny \pm 1.44}$ (40.48) & $\mathbf{98.98_{\tiny \pm 0.20} (1.00)}$  & $67.15_{\tiny \pm 1.05}$ (7.27) & $92.85_{\tiny \pm 0.97}$ (41.95) & 4.5 \\
&  &  & Default+TA & $\mathbf{35.23_{\tiny \pm 2.74} (12.55)}$ & $95.93_{\tiny \pm 0.50}$ (4.00) & $\mathbf{70.79_{\tiny \pm 0.88} (6.12)}$ & $\mathbf{68.17_{\tiny \pm 1.14} (23.28)}$ & 4.8 \\
\cline{3-9}
&  & \multirow{2}{*}{FT} & Default & $2.62_{\tiny \pm 0.17}$ (22.70) & $\mathbf{99.93_{\tiny \pm 0.01} (0.05)}$ & $\mathbf{75.56_{\tiny \pm 0.21} (1.14)}$ & $12.30_{\tiny \pm 0.35}$ (38.60) & 4.6 \\
&  &  & Default+TA & $\mathbf{9.88_{\tiny \pm 0.71} (12.79)}$ & $97.23_{\tiny \pm 0.23}$ (2.70) & $74.49_{\tiny \pm 0.61}$ (2.42) & $\mathbf{21.40_{\tiny \pm 1.15} (23.48)}$ & 4.1 \\
\cline{2-9}
  & \multirow{8}{*}{50\%} & \multirow{2}{*}{Retrain} & Default & $32.80_{\tiny \pm 0.28}$ (0.00) & $99.99_{\tiny \pm 0.01}$ (0.00) & $67.37_{\tiny \pm 0.53}$ (0.00) & $60.77_{\tiny \pm 0.57}$ (0.00) & 31.2 \\
  &  &  & Default+TA & $29.14_{\tiny \pm 0.31}$ (0.00) &$99.97_{\tiny \pm 0.01}$ (0.00)  & $70.77_{\tiny \pm 0.18}$ (0.00) & $54.37_{\tiny \pm 0.59}$ (0.00) & 31.3 \\
  \cline{3-9}
&  & \multirow{2}{*}{SalUn} & Default & $60.07_{\tiny \pm 2.82}$ (27.27) & $\mathbf{95.65_{\tiny \pm 0.89} (4.34)}$ &$46.61_{\tiny \pm 1.80}$ (20.76)  & $82.84_{\tiny \pm 0.66}$ (22.07) & 4.6 \\
&  &  & Default+TA & $\mathbf{39.58_{\tiny \pm 2.23} (10.44)}$ & $92.38_{\tiny \pm 0.74}$ (7.58) & $\mathbf{55.47_{\tiny \pm 1.41} (15.30)}$ & $\mathbf{66.33_{\tiny \pm 1.01} (11.96)}$ & 5.4 \\
\cline{3-9}
&  & \multirow{2}{*}{RL} & Default & $67.83_{\tiny \pm 2.26}$ (35.02) & $\mathbf{96.34_{\tiny \pm 0.55} (3.65)}$ & $45.18_{\tiny \pm 1.59}$ (22.19) & $87.76_{\tiny \pm 0.31}$ (26.99) & 4.5 \\
&  &  & Default+TA &  $\mathbf{44.96_{\tiny \pm 2.12} (15.81)}$ & $92.57_{\tiny \pm 1.01}$ (7.40) & $\mathbf{55.06_{\tiny \pm 1.45} (15.71)}$ & $\mathbf{71.52_{\tiny \pm 1.12} (17.15)}$ &5.0 \\
\cline{3-9}
&  & \multirow{2}{*}{FT} & Default & $2.63_{\tiny \pm 0.11}$ (30.17) & $\mathbf{99.97_{\tiny \pm 0.01} (0.02)}$ & $75.37_{\tiny \pm 0.22}$ (8.00) & $12.23_{\tiny \pm 0.16}$ (48.54) & 1.9 \\
&  &  & Default+TA & $\mathbf{8.25_{\tiny \pm 0.42} (20.90)}$ &$98.44_{\tiny \pm 0.19}$ (1.52)  &$\mathbf{73.9_{\tiny \pm 0.31} (3.22)}$  & $\mathbf{23.92_{\tiny \pm 0.67} (30.44)}$ & 2.3 \\

\bottomrule
\end{tabular}
}
\end{table*}

We evaluated the SOTA machine unlearning methods SalUn~\cite{fan2024salun}, Random Labeling (RL)~\cite{randomlabel} and Fine-Tuning (FT)~\cite{finetune} on the CIFAR-10 and CIFAR-100 datasets. These methods are assessed using random forget rates of 10\% and 50\%, simulating varying degrees of data removal requests. The performance of each method is compared to an ideal retrain scenario, which serves as the baseline, by measuring the Average Gap (AG). AG is computed as the mean of performance differences across the metrics Unlearn Accuracy (UA), Remaining Accuracy (RA), Testing Accuracy (TA), and Membership Inference Attack (MIA). Lower AG values indicate better performance. We adopt the Average Gap (AG) as a summary metric because it consolidates differences across multiple performance indicators (UA, RA, TA, and MIA) into a single value. This facilitates comparison across methods and augmentation strategies, providing a balanced view of unlearning quality while avoiding reliance on a single metric. We evaluated the unlearning process using 7 augmentation scenarios, described in Sec.~\ref{sec:metrics}: NoAug, Default (Random + Crop), Default + RandAugment, Default + AutoAugment, Default + Random Eraser, Default + TrivialAug and Default + AugMix. Figure~\ref{fig:gap_cifar10} shows the AG results for CIFAR-10, under 10\% and 50\% random forget rates, evaluating each unlearning method with various augmentation strategies. 
The absence of augmentation (NoAug) consistently results in higher AG values, indicating diminished unlearning effectiveness. The Fine-Tuning method significantly benefits from augmentation, especially with TrivialAug, showing notable AG reductions. However, SalUn and Random Labeling methods demonstrate relatively lower sensitivity to augmentation variations, primarily due to their inherently low baseline AG values.

Figure~\ref{fig:gap_cifar100} presents the AG results for random forgetting for CIFAR-100 dataset, with forget rates of 10\% and 50\%. For CIFAR-100, we can also observe that using no augmentation increases the AG. As the default AG values are high in the default configuration, we can observe a higher reduction in AG when using different augmentation strategies, for SalUn, Random Labeling and Fine-Tuning. Results show that Default + TrivialAug obtained the best results, achieving a 40.12\% and 18.61\% reduction in SalUn AG for 10\% and 50\% forget rates, respectively. This indicates that in challenging unlearning scenarios, such as CIFAR-100, data augmentation plays an important role in the unlearning process. Improvements in AG could also be observed in Random Label and Fine-Tuning when different data augmentation methods are used.

 Fine-Tuning benefits the most from augmentation, since it directly trains the model only on the reduced dataset $D_r$. When $D_r$ is small, especially under high forget rates, Fine-Tuning tends to overfit. Augmentation provides additional diversity that alleviates this overfitting, substantially improving generalisation and narrowing the performance gap to retrained models. In contrast, SalUn shows smaller gains because its parameter updates are also guided by saliency maps rather than only data diversity, making it inherently less dependent on augmentation. 
Another important observation is the dataset-specific effect. CIFAR-100 consistently exhibits larger performance gaps compared to CIFAR-10. This is expected, given the higher complexity of CIFAR-100, which includes 100 fine-grained classes with fewer samples per class.
 Such conditions increase the risk of overfitting and memorisation, making the unlearning task inherently more challenging. 

Table~\ref{tab:metrics} details the individual metrics of UA, RA, TA, MIA and RTE for the methods SalUn, RL and FT, comparing the default augmentation and Default+TrivialAug (Default+TA), which previously demonstrated the best AG performance, as shown in Figure~\ref{fig:gap_cifar10} and Figure~\ref{fig:gap_cifar100}. The gap of each metric compared to the retrain is shown in parentheses. The values with lower gaps are shown in bold. Results show that for CIFAR-10, most of the results remain close to the retrain performance, with a low gap of each metric using default or default+TA augmentation. For CIFAR-100, we observed a higher improvement in each individual metric gap, with an increased Test Accuracy metric and MIA for most of the cases. 
Notably, augmentation strategies do not significantly affect runtime efficiency, maintaining considerable speed advantages compared to full retraining.

\begin{figure*}[htp]
    \centering
    \includegraphics[width=0.85\textwidth]{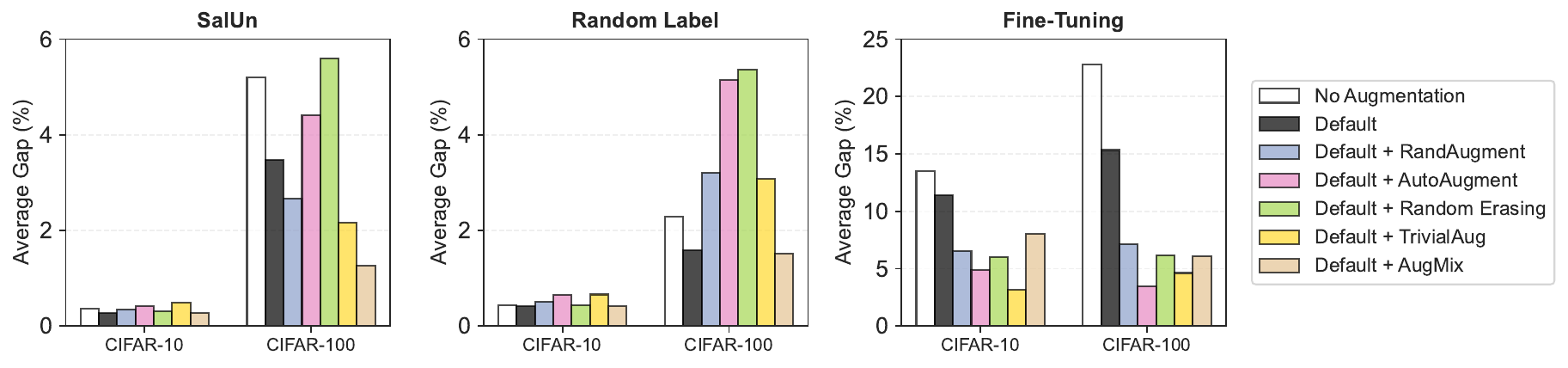}
    \caption{Average Gap (AG) under class-wise forgetting on CIFAR-10 and CIFAR-100. Results highlight the sensitivity of augmentation choices, with TrivialAug and AugMix offering notable benefits in specific methods and datasets.}
    \label{fig:gap_class}
\end{figure*}

We also evaluated the AG of the class-wise forgetting scenario. In this scenario, all the samples from a random class are unlearned. Figure~\ref{fig:gap_class} shows the AG results of the unlearning methods on the CIFAR-10 and CIFAR-100 datasets. For the class-wise forgetting scenario, we notice an improvement in the Fine-Tuning method, when using different strategies, with TrivilAug achieving the highest reduction. For SalUn, the reduction was higher for CIFAR-100, using Augmix. For Random Label, the use of data augmentations did not show improvements, and in some cases, it can increase the gap. Results show that class-wise unlearning is a challenging scenario, which requires a more careful selection of data augmentation.

From the results presented, we can conclude that compared to not using augmentation, the augmented strategies decrease the gap between the unlearned and retrained models. We also could observe higher benefits of using augmentation in challenging scenarios, as for CIFAR-100 with 50\% forget rate. We believe  that data augmentation is beneficial because it reduces data memorization, which harms the unlearning process. Also, we show that optimizing the augmentation strategies can further reduce the gap of unlearned models, and should be included in the MU design.

\section{Conclusion}

This work provides the first systematic study of data augmentation in MU. Experiments on CIFAR-10/100 with SalUn, Random Label, and Fine-Tuning show that carefully chosen augmentations reduce the gap to retrained models. We also highlight dataset and method-specific effects, as well as the role of augmentation in class-wise forgetting. These findings suggest that augmentation should be integrated into MU design, and future work should extend our analysis to larger datasets and architectures.

 Future work should explore larger and more diverse datasets, such as ImageNet, and alternative architectures, including Vision Transformers (ViTs) and more recent convolutional models. Such extensions would provide a broader validation of the role of augmentation in unlearning, further strengthening the generalizability of our findings.

\bibliographystyle{IEEEtran}
\bibliography{main}

\begin{thebibliography}{10}
\providecommand{\url}[1]{#1}
\csname url@samestyle\endcsname
\providecommand{\newblock}{\relax}
\providecommand{\bibinfo}[2]{#2}
\providecommand{\BIBentrySTDinterwordspacing}{\spaceskip=0pt\relax}
\providecommand{\BIBentryALTinterwordstretchfactor}{4}
\providecommand{\BIBentryALTinterwordspacing}{\spaceskip=\fontdimen2\font plus
\BIBentryALTinterwordstretchfactor\fontdimen3\font minus \fontdimen4\font\relax}
\providecommand{\BIBforeignlanguage}[2]{{%
\expandafter\ifx\csname l@#1\endcsname\relax
\typeout{** WARNING: IEEEtran.bst: No hyphenation pattern has been}%
\typeout{** loaded for the language `#1'. Using the pattern for}%
\typeout{** the default language instead.}%
\else
\language=\csname l@#1\endcsname
\fi
#2}}
\providecommand{\BIBdecl}{\relax}
\BIBdecl

\bibitem{buick2025copyright}
A.~Buick, ``Copyright and ai training data—transparency to the rescue?'' \emph{Journal of Intellectual Property Law and Practice}, vol.~20, no.~3, pp. 182--192, 2025.

\bibitem{hoofnagle2019european}
C.~J. Hoofnagle, B.~Van Der~Sloot, and F.~Z. Borgesius, ``The european union general data protection regulation: what it is and what it means,'' \emph{Information \& Communications Technology Law}, vol.~28, no.~1, pp. 65--98, 2019.

\bibitem{dang2021right}
Q.-V. Dang, ``Right to be forgotten in the age of machine learning,'' in \emph{Advances in Digital Science: ICADS 2021}.\hskip 1em plus 0.5em minus 0.4em\relax Springer, 2021, pp. 403--411.

\bibitem{mantelero2013eu}
A.~Mantelero, ``The eu proposal for a general data protection regulation and the roots of the ‘right to be forgotten’,'' \emph{Computer Law \& Security Review}, vol.~29, no.~3, pp. 229--235, 2013.

\bibitem{jaar2008canadian}
D.~Jaar and P.~E. Zeller, ``Canadian privacy law: The personal information protection and electronic documents act (pipeda),'' \emph{Int'l. In-House Counsel J.}, vol.~2, p. 1135, 2008.

\bibitem{randomlabel}
A.~Golatkar, A.~Achille, and S.~Soatto, ``Eternal sunshine of the spotless net: Selective forgetting in deep networks,'' in \emph{Proceedings of the IEEE/CVF conference on computer vision and pattern recognition}, 2020, pp. 9304--9312.

\bibitem{graves2021amnesiac}
L.~Graves, V.~Nagisetty, and V.~Ganesh, ``Amnesiac machine learning,'' in \emph{Proceedings of the AAAI Conference on Artificial Intelligence}, vol.~35, no.~13, 2021, pp. 11\,516--11\,524.

\bibitem{shanahan2024talking}
M.~Shanahan, ``Talking about large language models,'' \emph{Communications of the ACM}, vol.~67, no.~2, pp. 68--79, 2024.

\bibitem{zhang2023review}
H.~Zhang, T.~Nakamura, T.~Isohara, and K.~Sakurai, ``A review on machine unlearning,'' \emph{SN Computer Science}, vol.~4, no.~4, p. 337, 2023.

\bibitem{di2024label}
Z.~Di, Z.~Zhu, J.~Jia, J.~Liu, Z.~Takhirov, B.~Jiang, Y.~Yao, S.~Liu, and Y.~Liu, ``Label smoothing improves machine unlearning,'' \emph{arXiv preprint arXiv:2406.07698}, 2024.

\bibitem{jia2023model}
J.~Jia, J.~Liu, P.~Ram, Y.~Yao, G.~Liu, Y.~Liu, P.~Sharma, and S.~Liu, ``Model sparsity can simplify machine unlearning,'' \emph{Advances in Neural Information Processing Systems}, vol.~36, pp. 51\,584--51\,605, 2023.

\bibitem{zhao2024makes}
K.~Zhao, M.~Kurmanji, G.-O. B{\u{a}}rbulescu, E.~Triantafillou, and P.~Triantafillou, ``What makes unlearning hard and what to do about it,'' \emph{Advances in Neural Information Processing Systems}, vol.~37, pp. 12\,293--12\,333, 2024.

\bibitem{mumuni2024survey}
A.~Mumuni, F.~Mumuni, and N.~K. Gerrar, ``A survey of synthetic data augmentation methods in machine vision,'' \emph{Machine Intelligence Research}, vol.~21, no.~5, pp. 831--869, 2024.

\bibitem{fan2024salun}
\BIBentryALTinterwordspacing
C.~Fan, J.~Liu, Y.~Zhang, E.~Wong, D.~Wei, and S.~Liu, ``Salun: Empowering machine unlearning via gradient-based weight saliency in both image classification and generation,'' in \emph{International Conference on Learning Representations (ICLR)}, 2024. [Online]. Available: \url{https://arxiv.org/abs/2310.12508}
\BIBentrySTDinterwordspacing

\bibitem{finetune}
A.~Warnecke, L.~Pirch, C.~Wressnegger, and K.~Rieck, ``Machine unlearning of features and labels,'' \emph{arXiv preprint arXiv:2108.11577}, 2021.

\bibitem{cifar}
\BIBentryALTinterwordspacing
A.~Krizhevsky, ``Learning multiple layers of features from tiny images,'' University of Toronto, Tech. Rep., 2009. [Online]. Available: \url{https://www.cs.toronto.edu/~kriz/learning-features-2009-TR.pdf}
\BIBentrySTDinterwordspacing

\bibitem{li2024taxonomy}
\BIBentryALTinterwordspacing
N.~Li, C.~Zhou, Y.~Gao, H.~Chen, A.~Fu, Z.~Zhang, and S.~Yu, ``Machine unlearning: Taxonomy, metrics, applications, challenges, and prospects,'' \emph{ACM Computing Surveys}, 2024. [Online]. Available: \url{https://doi.org/10.1145/nnnnnnn.nnnnnnn}
\BIBentrySTDinterwordspacing

\bibitem{bourtoule2021machine}
L.~Bourtoule, V.~Chandrasekaran, C.~A. Choquette-Choo, H.~Jia, A.~Travers, B.~Zhang, D.~Lie, and N.~Papernot, ``Machine unlearning,'' in \emph{2021 IEEE symposium on security and privacy (SP)}.\hskip 1em plus 0.5em minus 0.4em\relax IEEE, 2021, pp. 141--159.

\bibitem{golatkar2020eternal}
A.~Golatkar, A.~Achille, and S.~Soatto, ``Eternal sunshine of the spotless net: Selective forgetting in deep networks,'' in \emph{Proceedings of the IEEE/CVF conference on computer vision and pattern recognition}, 2020, pp. 9304--9312.

\bibitem{warnecke2021machine}
A.~Warnecke, L.~Pirch, C.~Wressnegger, and K.~Rieck, ``Machine unlearning of features and labels,'' \emph{arXiv preprint arXiv:2108.11577}, 2021.

\bibitem{takahashi2019data}
R.~Takahashi, T.~Matsubara, and K.~Uehara, ``Data augmentation using random image cropping and patching for deep cnns,'' \emph{IEEE Transactions on Circuits and Systems for Video Technology}, vol.~30, no.~9, pp. 2917--2931, 2019.

\bibitem{randaug}
E.~D. Cubuk, B.~Zoph, J.~Shlens, and Q.~V. Le, ``Randaugment: Practical automated data augmentation with a reduced search space,'' in \emph{Proceedings of the IEEE/CVF conference on computer vision and pattern recognition workshops}, 2020, pp. 702--703.

\bibitem{autoaug}
E.~D. Cubuk, B.~Zoph, D.~Mane, V.~Vasudevan, and Q.~V. Le, ``Autoaugment: Learning augmentation policies from data,'' \emph{arXiv preprint arXiv:1805.09501}, 2018.

\bibitem{randomerasing}
Z.~Zhong, L.~Zheng, G.~Kang, S.~Li, and Y.~Yang, ``Random erasing data augmentation,'' in \emph{Proceedings of the AAAI conference on artificial intelligence}, vol.~34, no.~07, 2020, pp. 13\,001--13\,008.

\bibitem{trivial}
S.~G. M{\"u}ller and F.~Hutter, ``Trivialaugment: Tuning-free yet state-of-the-art data augmentation,'' in \emph{Proceedings of the IEEE/CVF international conference on computer vision}, 2021, pp. 774--782.

\bibitem{augmix}
D.~Hendrycks, N.~Mu, E.~D. Cubuk, B.~Zoph, J.~Gilmer, and B.~Lakshminarayanan, ``{AugMix}: A simple data processing method to improve robustness and uncertainty,'' \emph{Proceedings of the International Conference on Learning Representations (ICLR)}, 2020.

\bibitem{wu2019wider}
Z.~Wu, C.~Shen, and A.~Van Den~Hengel, ``Wider or deeper: Revisiting the resnet model for visual recognition,'' \emph{Pattern recognition}, vol.~90, pp. 119--133, 2019.

\end{thebibliography}
%

\end{document}